\documentclass[11pt]{article}
\usepackage{acl2012}
\usepackage{times}
\usepackage{latexsym}
\usepackage{amsmath}
\usepackage{url}

\usepackage{amsmath}
\usepackage{amssymb}
\usepackage{amsthm}

 \author{Daoud Clarke \\
 Gorkana Group\\
 28--42 Banner Street, London\\
  {\tt daoud.clarke@gorkana.com} }

\date{\today}

\title{Challenges for Distributional Compositional Semantics}

\begin{document}

\maketitle

\begin{abstract}
This paper summarises the current state-of-the art in the study of
compositionality in distributional semantics, and major challenges for
this area. We single out generalised quantifiers and intensional
semantics as areas on which to focus attention for the development of
the theory. Once suitable theories have been developed, algorithms
will be needed to apply the theory to tasks. Evaluation is a major
problem; we single out application to recognising textual entailment and
machine translation for this purpose.
\end{abstract}

\section{Introduction}

This paper summarises some major challenges for the nascent field of
distributional compositional semantics. Research in this area has
arisen out of the success of vector-based techniques for representing
aspects of lexical semantics, such as latent semantic analysis
\cite{Deerwester:90} and measures of distributional similarity
\cite{Lin:98,Lee:99}.

The automatic nature of these techniques mean that much higher
coverage can be achieved compared to manually constructed resources
such as WordNet \cite{Fellbaum:05}. Additionally, the vector-based
nature of the semantic representations allow for fine-grained aspects
of meaning to be incorporated, in contrast to the type of relations
typically expressed in ontologies; moreover the construction of an
ontology is generally a subjective process, whereas vector-based
approaches are typically more objective, being formed from
observations of the contexts in which words occur in large
corpora. There are disadvantages: automatic techniques are arguably
less reliable than manually constructed resources, and often do not
explicitly identify the variety of relationships between words that
are captured in an ontology such as WordNet.

Researchers have begun to look at how such techniques can be extended
beyond the word level to represent meanings of phrases and even whole
sentences. Existing techniques cannot be applied directly beyond
phrases of two or three words because of the problem of data
sparseness --- as the length of the phrase increases, the amount
data matching the phrase falls off very quickly, and soon there is not
enough data to build vectors reliably. The alternative is to look at
how to compose such vectors, so that the vector for a phrase or
sentence is determined purely by the vector representations for the
individual words in the sentence.

While interest in this area has exploded in recent years, and some
significant advances have been made, there is still a lot of work to
do:
\begin{itemize}
\item The underlying theory needs to be developed to allow
  distributional approaches to describe aspects of natural language
  meaning easily described by model-theoretic semantics, for example,
  generalised quantifiers and intensional semantics. We explain below
  why current approaches are not suited to either of these.
\item New algorithms and tools are needed to perform inference with
  the new theories.
\item We need suitable methods for evaluating distributional models of
  compositionality. In addition, approaches need to be evaluated
  across a broader range of natural language processing tasks. In
  particular we identify textual entailment and machine translation as
  suitable areas for application of current and future techniques.
\end{itemize}

In the remainder of the paper, we summarise existing work (Section
\ref{background}), then motivate each of the above areas in detail
(Section \ref{challenges}).

\section{Background}
\label{background}

Vector representations provide a rich variety of possible methods of
composition. The most obvious method is perhaps vector addition
\cite{Landauer:97,Foltz:98}, in which a string of words is represented
by the sum of the individual words making up the string. This method
has several problems, the most obvious of which is that the operation
is commutative, whereas natural language meaning is not: \emph{John
  hit Mary} does not mean the same as \emph{Mary hit John}. Another
composition operation that suffers from this problem is point-wise
multiplication \cite{Mitchell:08}.

A method of composing vectors that avoids this issue is the tensor
product \cite{Smolensky:90,Clark:07,Widdows:08}. Given two vectors $u$
and $v$ in vector spaces $U$ and $V$ of dimensionality $m$ and $n$
respectively, the tensor product $u\otimes v$ is a vector in a much
larger space $U\otimes V$ of dimensionality $mn$. Each pair of basis
vectors in $U$ and $V$ has a corresponding basis vector in $U\otimes
V$, so given a tensor product $u\otimes v$ it is always possible to
deduce the original vectors $u$ and $v$, another property that is
missing from vector addition.

The problem with the tensor product is that strings of different
lengths have different dimensionalities and live in different vector
spaces and are thus not directly comparable. This means that we cannot
say to what extent \emph{big dog} entails \emph{dog}. There are
several ways to get around this:
\begin{itemize}
\item Use some linear map from the tensor product space to the
  original space to reduce the dimensionality of vectors and allow
  them to be compared. This was suggested by \newcite{Mitchell:08} as
  a general ``multiplicative model'' of composition. The problem with
  this method is that information is lost as meanings compose since
  all strings have the same dimensionality.
\item Impose relations on different tensor powers of the space to make
  them comparable \cite{Clarke:10}. This approach allows a lot of
  flexibility in describing composition but it is not clear how to
  determine what relations should be imposed, nor how we can easily
  compute with the resulting structures. It does, however, resolve the
  problem of information loss as strings are composed.
\end{itemize}


The approach of \newcite{Grefenstette:11} is inspired by some
mathematical similarities between the structure of vector spaces and
that of pregroup grammars: they are both compact closed
categories. Their approach can be viewed as a vectorisation of
Montague semantics \cite{Clark:08}.

Other approaches to this problem include the use of matrices
\cite{Rudolph:10} including those learnt directly from data
\cite{Baroni:10}.



\subsection{Context-theoretic Semantics}

The framework of \newcite{Clarke:12} is a mathematical formalisation
of the idea that meaning is determined by context. The structure that
is proposed to model natural language semantics is an associative
algebra over the real numbers $\mathbb{R}$. This is a real vector
space $\mathcal{A}$, together with multiplication which satisfies a
property called \textbf{bilinearity}:
\begin{align*}
a(\alpha b + \beta c) &= \alpha ab + \beta ac\\
(\alpha a+\beta b)c &= \alpha ac + \beta bc
\end{align*}
for all $a,b,c \in \mathcal{A}$ and all $\alpha, \beta \in
\mathbb{R}$. It can be shown that this type of structure generalises
all the approaches we discussed above \cite{Clarke:12}.

\newcite{Clarke:12} also proposes a principle to determine entailment
between strings in distributional semantics, based on the concept
of \textbf{distributional generality} \cite{Weeds:04}, that terms that
have a more general meaning will occur in a wider range of
contexts. The theory assumes the existence of a distinguished basis
which can be interpreted as defining the contexts in which strings can
appear. This defines a partial ordering on the vector space by $u \le
v$ if and only if every component of $u$ is less than or equal to the
corresponding component of $v$. The partial ordering is interpreted as
entailment and is connected with distributional generality since
$\hat{x} \le \hat{y}$ if $y$ occurs at least as frequently as $x$ in
every context, where $\hat{x}$ and $\hat{y}$ are the vectors
associated with terms $x$ and $y$. 


\section{Challenges}
\label{challenges}

\subsection{Theory}

The greatest problem currently facing attempts to describe meaning
using vectors is to reconcile them with existing theories of meaning,
most notably logical approaches to semantics. If distributional
semantics is to replace logical semantics, it has to encompass it,
since there are things that logical semantics does very well that it
is hard to imagine distributional semantics doing in its current
form. For example, it is conceivable that an intelligent agent could
be built which interpreted natural language sentences using logic. The
agent would chose the best course of action given a set of
assumptions, perhaps using a combination of theorem provers, automated
planning and search tools. The functionality provided by the theorem
proving component in such a system would be essential, allowing
diverse pieces of knowledge from a variety of sources to be combined
and deductions to be made from them. This is something that
distributional approaches are not currently able to do.

Encompassing a whole logical semantic formalism in a manner
consistent with distributional semantics is an ambitious goal. We have
identified two particular areas with the following characteristics:
\begin{itemize}
\item They are intuitively familiar and easy to understand
\item They occur fairly frequently in ordinary speech and writing
\item No existing framework for compositionality in distributional
  semantics deals with them satisfactorily
\end{itemize}
It is our hope that by concentrating on these areas we are able to make
progress towards the ultimate goal.

\subsubsection*{Generalised Quantifiers}

The study of generalised quantifiers concerns expressions such as
\emph{some}, \emph{most but not all}, \emph{no} and \emph{at least
  two}. In the analysis of \newcite{Barwise:81}, which is based on the
earlier work of \newcite{Montague:74}, the semantics of determiners
such as these is to operate on a set of entities (for example the set
of people) and to return a set of sets, for example the semantics of
\emph{most people} is the set of all sets of entities which contain
most people.

Formalising these properties mathematically allows us to understand
some properties of entailment between sentences containing such
quantifiers. For example, \emph{all animals breathe} entails \emph{all
  cats breathe}, whereas \emph{some cats like cheese} entails \emph{some
  animals like cheese}; the change in quantifier has reversed the
direction of the entailment.

This property cannot be captured within the framework of
\newcite{Clarke:12}, because of the in-built property of linearity of
the multiplication in the underlying algebra. If we accept the idea of
distributional generality, that \emph{cat} should entail \emph{animal}
because the latter will occur in a broader range of contexts, then it
follows from linearity that \emph{x cat y} will entail \emph{x animal
  y} for any strings \emph{x} and \emph{y}. More generally, for any
$u,v,w\in \mathcal{A}$ such that $u\le v$, $uw \le vw$ and $wu \le
wv$.

In fact, what the reversal of entailment indicates is that quantifiers
such as \emph{all} are \textbf{non-linear}; they are not compatible
with the bilinearity condition of context-theoretic semantics. This is
a problem for all existing approaches to the problem of
compositionality in distributional semantics, since linearity is a
common assumption among them \cite{Clarke:12}.

The work of \newcite{Preller:09} addresses the problem of representing
negation in distributional semantics using Bell states. Since negation
results in a similar reversal of entailment, it is possible that such
an approach would also be useful for modelling generalised
quantifiers.

\subsubsection*{Intensional Semantics}

Intensional semantics deals with certain complex semantic phenomena
such as those involving the verbs \emph{know}, \emph{believe},
\emph{want} and \emph{need}. These are described elegantly in Montague
semantics \cite{Montague:74}, and the ability to reason about such
concepts is essential for intelligent agents that would interact with
humans in natural language. Reasoning about such sentences requires
additional knowledge about the meaning of these words that would
normally be described in terms of logic; it is hard to imagine how
their meanings could be described reliably within distributional
semantics.

\subsection{Algorithms and Tools}

In order to compete with logical methods in semantics, distributional
semantics needs to be able to, given a fixed set of background
knowledge (expressed in natural language):
\begin{enumerate}
\item \textbf{Truth:} Estimate the probability that a given sentence is true.
\item \textbf{Search:} Given a parameterised sentence, for example
  \emph{the queen was born in $x$}, find the parameter $x$ which
  maximises the probability of the sentence.
\item \textbf{Entailment:} Given two sentences, compute the degree to
  which the first entails the second.
\end{enumerate}
The first and third of these will be useful in tasks such as question
answering while the third will be useful for any of the tasks
associated with textual entailment \cite{Dagan:05}, for example
information retrieval.

There are more complex tasks that may not be expressible in terms of
distributional semantics, for example those needed in planning for an
intelligent agent; the exact formulation for such tasks may depend on
the the particular semantic formalism chosen.

When designing algorithms for these tasks, it is likely that we will
be able to compute the answer much faster if we allow an approximation
to the answer, which may be perfectly suitable for many tasks. Without
a satisfactory theory of meaning, however, it is hard to speculate on the
possible nature of such algorithms.

\subsection{Evaluation Methods}

A problem for researchers working in this field is how to evaluate
models of compositionality. Researchers have evaluated models on short
phrases by determining context vectors for the phrases and for
individual words directly. They then compose the vectors for
individual words using their models to obtain vectors for phrases and
measure how similar these are to the observed phrase vectors
\cite{Baroni:10,Guevara:11}. This evaluation technique cannot be
extended beyond short phrases however, so may not provide a good
measure of how good models are at handling deep semantics.

The recent Workshop on Distributional Semantics and Compositionality
\cite{Biemann:11} provided a dataset and a shared task of determining
to what degree a phrase is compositional. This is undoubtedly a useful
task, but again does not address the question of deep semantics.

In order to evaluate deep semantics, we propose applying methods to
two tasks requiring deep semantics to perform well: recognising
textual entailment and machine translation. We believe these tasks are
suitable for this purpose because they would intuitively seem to
require deep semantics to achieve perfect performance, yet statistical
approaches are able to achieve reasonable to good performance. These
tasks would thus provide a testing ground in which the sophistication
of the techniques applied can be increased gradually towards deep
semantics, the hope being that the more sophisticated techniques will
lead to improved performance.

\section{Conclusion}

We have summarised some approaches to modelling compositionality in
distributional semantics, and highlighted some challenges which we
believe to be pertinent. In particular, we identified some aspects of
the theory of distributional semantics which we believe to be lacking;
anyone able to resolve these will necessarily push the boundaries of
our understanding of meaning.

\bibliographystyle{acl2012}
\bibliography{contexts}

\begin{thebibliography}{}

\bibitem[\protect\citename{Baroni and Zamparelli}2010]{Baroni:10}
Marco Baroni and Roberto Zamparelli.
\newblock 2010.
\newblock Nouns are vectors, adjectives are matrices: Representing
  adjective-noun constructions in semantic space.
\newblock In {\em Proceedings of the Conference on Empirical Methods in Natural
  Language Processing (EMNLP 2010), East Stroudsburg PA: ACL}, pages
  1183--1193.

\bibitem[\protect\citename{Barwise and Cooper}1981]{Barwise:81}
Jon Barwise and Robin Cooper.
\newblock 1981.
\newblock Generalized quantifiers and natural language.
\newblock {\em Linguistics and Philosophy}, 4:159--219.

\bibitem[\protect\citename{Biemann and Giesbrecht}2011]{Biemann:11}
Chris Biemann and Eugenie Giesbrecht, editors.
\newblock 2011.
\newblock {\em Proceedings of the Workshop on Distributional Semantics and
  Compositionality}.
\newblock Association for Computational Linguistics, Portland, Oregon, USA,
  June.

\bibitem[\protect\citename{Clark and Pulman}2007]{Clark:07}
Stephen Clark and Stephen Pulman.
\newblock 2007.
\newblock Combining symbolic and distributional models of meaning.
\newblock In {\em Proceedings of the AAAI Spring Symposium on Quantum
  Interaction}, pages 52--55, Stanford, CA.

\bibitem[\protect\citename{Clark \bgroup et al.\egroup }2008]{Clark:08}
Stephen Clark, Bob Coecke, and Mehrnoosh Sadrzadeh.
\newblock 2008.
\newblock A compositional distributional model of meaning.
\newblock In {\em Proceedings of the Second Quantum Interaction Symposium
  (QI-2008)}, pages 133--140, Oxford, UK.

\bibitem[\protect\citename{Clarke \bgroup et al.\egroup }2010]{Clarke:10}
Daoud Clarke, Rudi Lutz, and David Weir.
\newblock 2010.
\newblock Semantic composition with quotient algebras.
\newblock In {\em Proceedings of the 2010 Workshop on GEometrical Models of
  Natural Language Semantics}, pages 38--44, Uppsala, Sweden, July. Association
  for Computational Linguistics.

\bibitem[\protect\citename{Clarke}2012]{Clarke:12}
Daoud Clarke.
\newblock 2012.
\newblock A context-theoretic framework for compositionality in distributional
  semantics.
\newblock {\em Computational Linguistics}, 38(1):41--71.

\bibitem[\protect\citename{Dagan \bgroup et al.\egroup }2005]{Dagan:05}
Ido Dagan, Oren Glickman, and Bernardo Magnini.
\newblock 2005.
\newblock The pascal recognising textual entailment challenge.
\newblock In {\em Proceedings of the PASCAL Challenges Workshop on Recognising
  Textual Entailment}, pages 1--8.

\bibitem[\protect\citename{Deerwester \bgroup et al.\egroup
  }1990]{Deerwester:90}
Scott Deerwester, Susan Dumais, George Furnas, Thomas Landauer, and Richard
  Harshman.
\newblock 1990.
\newblock Indexing by latent semantic analysis.
\newblock {\em Journal of the American Society for Information Science},
  41(6):391--407.

\bibitem[\protect\citename{Fellbaum}2005]{Fellbaum:05}
C.~Fellbaum.
\newblock 2005.
\newblock Wordnet and wordnets.
\newblock {\em Encyclopedia of Language and Linguistics, Second Edition,
  Oxford: Elsevier}, pages 665--670.

\bibitem[\protect\citename{Foltz \bgroup et al.\egroup }1998]{Foltz:98}
Peter~W. Foltz, Walter Kintsch, and Thomas~K. Landauer.
\newblock 1998.
\newblock The measurement of textual coherence with latent semantic analysis.
\newblock {\em Discourse Process}, 15:285--307.

\bibitem[\protect\citename{Grefenstette \bgroup et al.\egroup
  }2011]{Grefenstette:11}
Edward Grefenstette, Mehrnoosh Sadrzadeh, Stephen Clark, Bob Coecke, and
  Stephen Pulman.
\newblock 2011.
\newblock Concrete sentence spaces for compositional distributional models of
  meaning.
\newblock {\em Proceedings of the 9th International Conference on Computational
  Semantics (IWCS 2011)}, pages 125--134.

\bibitem[\protect\citename{Guevara}2011]{Guevara:11}
Emiliano Guevara.
\newblock 2011.
\newblock Computing semantic compositionality in distributional semantics.
\newblock In {\em Proceedings of the 9th International Conference on
  Computational Semantics (IWCS 2011)}, pages 135--144.

\bibitem[\protect\citename{Landauer and Dumais}1997]{Landauer:97}
Thomas~K. Landauer and Susan~T. Dumais.
\newblock 1997.
\newblock A solution to {P}lato's problem: the latent semantic analysis theory
  of acquisition, induction and representation of knowledge.
\newblock {\em Psychological Review}, 104(2):211--240.

\bibitem[\protect\citename{Lee}1999]{Lee:99}
Lillian Lee.
\newblock 1999.
\newblock Measures of distributional similarity.
\newblock In {\em Proceedings of the 37th Annual Meeting of the Association for
  Computational Linguistics (ACL-1999)}, pages 23--32.

\bibitem[\protect\citename{Lin}1998]{Lin:98}
Dekang Lin.
\newblock 1998.
\newblock Automatic retrieval and clustering of similar words.
\newblock In {\em Proceedings of the 36th Annual Meeting of the Association for
  Computational Linguistics and the 17th International Conference on
  Computational Linguistics (COLING-ACL '98)}, pages 768--774, Montreal.

\bibitem[\protect\citename{Mitchell and Lapata}2008]{Mitchell:08}
Jeff Mitchell and Mirella Lapata.
\newblock 2008.
\newblock Vector-based models of semantic composition.
\newblock In {\em Proceedings of ACL-08: HLT}, pages 236--244, Columbus, Ohio,
  June. Association for Computational Linguistics.

\bibitem[\protect\citename{Montague}1974]{Montague:74}
Richard Montague.
\newblock 1974.
\newblock The proper treatment of quantification in ordinary english.
\newblock In {\em Formal Philosophy: Selected Papers of Richard Montague}. Yale
  University Press.

\bibitem[\protect\citename{Preller and Sadrzadeh}2011]{Preller:09}
Anne Preller and Mehrnoosh Sadrzadeh.
\newblock 2011.
\newblock Bell states and negative sentences in the distributed model of
  meaning.
\newblock {\em Electronic Notes in Theoretical Computer Science},
  270(2):141--153.
\newblock Proceedings of the 6th International Workshop on Quantum Physics and
  Logic (QPL 2009).

\bibitem[\protect\citename{Rudolph and Giesbrecht}2010]{Rudolph:10}
Sebastian Rudolph and Eugenie Giesbrecht.
\newblock 2010.
\newblock Compositional matrix-space models of language.
\newblock In {\em Proceedings of the 48th Annual Meeting of the Association for
  Computational Linguistics}, pages 907--916, Uppsala, Sweden, July.
  Association for Computational Linguistics.

\bibitem[\protect\citename{Smolensky}1990]{Smolensky:90}
Paul Smolensky.
\newblock 1990.
\newblock Tensor product variable binding and the representation of symbolic
  structures in connectionist systems.
\newblock {\em Artificial Intelligence}, 46(1-2):159--216, November.

\bibitem[\protect\citename{Weeds \bgroup et al.\egroup }2004]{Weeds:04}
Julie Weeds, David Weir, and Diana McCarthy.
\newblock 2004.
\newblock Characterising measures of lexical distributional similarity.
\newblock In {\em Proceedings of Coling 2004}, pages 1015--1021, Geneva,
  Switzerland, Aug 23--Aug 27. COLING.

\bibitem[\protect\citename{Widdows}2008]{Widdows:08}
Dominic Widdows.
\newblock 2008.
\newblock Semantic vector products: Some initial investigations.
\newblock In {\em Proceedings of the Second Symposium on Quantum Interaction,
  Oxford, UK}, pages 1--8.

\end{thebibliography}

\end{document}